\documentclass[lettersize,journal]{IEEEtran}

\usepackage{amsmath,amssymb,amsfonts}
\usepackage{algorithmic}
\usepackage{algorithm}
\usepackage{array}
\usepackage[caption=false,font=normalsize,labelfont=sf,textfont=sf]{subfig}
\usepackage{textcomp}
\usepackage{stfloats}
\usepackage{url}
\usepackage{verbatim}
\usepackage{graphicx}
\usepackage{subcaption}
\usepackage{cite}
\usepackage{makecell}

\usepackage{booktabs}
\usepackage{multirow}
\usepackage{tikz}
\usetikzlibrary{arrows.meta,positioning,shapes,fit}
\usepackage{svg}
\usepackage{listings}
\begin{document}

          	\title{DriveMCP: An Agentic AI framework for Advanced Driver Assistance System}

          \author{Farzad~Nadiri,~\IEEEmembership{Member,~IEEE,} Mehdi~Cina,~\IEEEmembership{Member,~IEEE,}
                  and~Ahmad~B.~Rad,~\IEEEmembership{Senior Member,~IEEE}%
                  \thanks{Manuscript received Month DD, YYYY; revised Month DD, YYYY. (Corresponding author: Ahmad B. Rad.)}%
                  \thanks{F. Nadiri, M. Cina and A. B. Rad are with the Department of Mechatronics Engineering, Simon Fraser University, Burnaby, BC, Canada (corresponding author e-mail: arad@sfu.ca).}%
                  }

                  \markboth{IEEE Transactions on Intelligent Vehicles,~Vol.~X, No.~X, Month~YYYY}%
                  {Nadiri and Rad: DriveMCP: DriveLM-Grounded Agentic Driver Assistance via MCP}

                  \maketitle

                  \begin{abstract}
This paper introduces an agentic AI driver-assistance framework that integrates perception, compliance reasoning, vehicle-state interpretation, and safety arbitration into a modular and auditable pipeline.
The architecture, referred to as \emph{DriveMCP}, incorporates a sensor-like perception stack alongside DriveLM as the vision-language front end to generate a graph-structured scene understanding (Graph Visual Question Answering) and language-grounded driving information.
Key compliance elements in \texttt{world\_state}, including posted speed limits and jurisdiction cues, are derived from DriveLM outputs through a structured parsing layer rather than being injected as simulator ground truth.
A stateful orchestration layer coordinates specialized experts exposed as Model Context Protocol (MCP) servers: (i) a Rules server that performs retrieval-augmented compliance reasoning over jurisdiction-specific traffic codes and sign conventions, (ii) a Weather server that estimates traction risk and contextual speed advisories, and (iii) an MCP-CAN server that surfaces Controller Area Network (CAN)/On-Board Diagnostics (OBD) telemetry and diagnostic context for health-aware risk shaping.
These outputs are fused to generate a structured decision that prompts a recommended course of action.
The outcome is then further filtered by a Responsibility-Sensitive Safety (RSS)-inspired guardrail that arbitrates \emph{speak} versus \emph{act} decisions under bounded online adaptation.
In CARLA simulation across multilingual, cross-border, and dynamic speed-limit scenarios, DriveMCP reduces traffic infractions and overspeed relative to the VLM-Direct, VLM-Direct+RAG, and VLM-Tools-NoArbiter baselines, while improving hazard response time and maintaining sub-second advisory latency.
We further quantify robustness under corrupted \texttt{world\_state} fields and show that CAN-aware context improves recommended targets and TTC at intervention under degraded braking envelopes.
\end{abstract}


                  \section{Introduction}
\IEEEPARstart{S}{afe} driving entails a balanced fusion of several attributes including skills, vigilance, proactive and reactive control, and sound decision making, along with a synchronized execution of these processes.
Failure in any of these components or their real-time interactions can lead to catastrophic accidents.
The possibility of incorporating all of these requirements in advanced driver-assistance systems (ADAS) and autonomous driving appeared far-fetched only a few years ago.
However, the impressive and rapid development of AI in the last few years has drastically changed that narrative and the realization of ADAS and autonomous driving systems with such capabilities is within reach.
Consider a situation in which a driver is operating a vehicle in an unfamiliar jurisdiction, but accompanied by a multilingual companion who is knowledgeable about various driving conditions, understands the rules of the road, and is capable of assisting the driver in safely navigating the route.
In this paper, we introduce a system that emulates this scenario.

The current implementations increasingly incorporate on-the-fly perception and planning components, yet they remain constrained by the difficulty of encoding diverse traffic rules, explaining decisions to human drivers, and above all adapting to changing conditions such as weather, road conditions, and jurisdiction boundaries.
From a broader perspective, motion planning and control have been extensively studied \cite{paden_tiv_2016}, and deep-learning-based driving pipelines continue to evolve rapidly \cite{grigorescu_jfr_2020, chen_tpami_2024}.
Recent multimodal large language models (MLLMs) and vision-language(-action) models extend this trajectory by enabling natural-language queries, human-oriented explanations, and high-level reasoning over a wide range of driving scenes \cite{drivegpt4, drivelm, vlp, vlaad, simlingo, vla_survey}.

Representative driving-oriented systems include DriveGPT4 \cite{drivegpt4}, which couples multi-frame visual inputs with language-grounded control prediction and interpretability, and retrieval-augmented approaches such as RAG-Driver \cite{ragdriver} that ground explanations in external demonstrations or documents.
DriveLM \cite{drivelm} introduces graph-structured visual question answering for driving and a baseline agent that jointly performs GVQA and end-to-end driving.

Despite these advances, \emph{monolithic} VLM-based driving assistants raise three practical concerns for deployment in intelligent vehicles.
First, compliance and updateability: traffic regulations vary by country, state, and even municipality, and they evolve over time.
Embedding such rules solely in model parameters or long prompts hinders verification and makes updates costly.
Second, traceability: safety cases often require reconstructing why an assistant suggested or performed an action, including which rules and observations were used.
Third, latency and robustness: a single large inference that performs perception, rule reasoning, and decision-making can be slow and may fail in out-of-distribution conditions, particularly when multiple factors (e.g., signage and rain and degraded vehicle health) must be considered simultaneously.

In this paper, we present DriveMCP that addresses all the aforementioned challenges.
The proposed system combines a driving-specialized VLM (DriveLM) \cite{drivelm} with explicit expert services connected through Model Context Protocol (MCP) \cite{mcp} and is coordinated by a stateful orchestration graph (LangGraph-style) \cite{langgraph}.
The goal is not to replace certified ADAS controllers with an LLM, but to build an assistance layer that improves compliance reasoning, explainability, and context awareness, while enforcing conservative safety constraints with RSS-inspired checks \cite{rss}.
DriveMCP is designed to complement established perception--planning--control modules; for instance, specialized lateral-control strategies (e.g., look-down control) remain applicable beneath an advisory layer \cite{nadiri_machines_lookdown_2025}.
\subsection{Design Requirements}
                  Based on common requirements for driver assistance (human-in-the-loop, auditability, real-time response), we employ the following design requirements to guide the architecture (Table~\ref{tab:req}).

\begin{table}[t]
                  \centering
                  \caption{Key design requirements and corresponding DriveMCP mechanisms.}
                  \label{tab:req}
                  \setlength{\tabcolsep}{5pt}
                  \begin{tabular}{p{0.30\columnwidth}p{0.60\columnwidth}}
                  \toprule
                  \textbf{Requirement} & \textbf{Mechanisms} \\
                  \midrule
                  Regulation updateability &
                  \begin{itemize}\setlength{\itemsep}{0pt}
                    \item Rules MCP server
                      \item External regulatory corpus retrieval
                        \item Jurisdiction tagging and rule-pack versioning
                        \end{itemize} \\

                        Explainability and traceability &
                        \begin{itemize}\setlength{\itemsep}{0pt}
                          \item Structured \texttt{world\_state}
                            \item Tool-call and evidence logs
                              \item Fused human-readable explanations
                              \end{itemize} \\

                              Real-time responsiveness &
                              \begin{itemize}\setlength{\itemsep}{0pt}
                                \item Selective tool routing
                                  \item Parallel MCP calls
                                    \item Query caching
                                    \end{itemize} \\

                                    Safety conservatism &
                                    \begin{itemize}\setlength{\itemsep}{0pt}
                                      \item RSS and TTC-based guardrails
                                        \item Speak-versus-act arbitration
                                          \item Conservative fallback under uncertainty
                                          \end{itemize} \\

                                          Vehicle-health awareness &
                                          \begin{itemize}\setlength{\itemsep}{0pt}
                                            \item MCP-CAN telemetry
                                              \item DTC-aware context modeling
                                                \item Degraded-mode advisories
                                                \end{itemize} \\
                                                \bottomrule
                                                \end{tabular}
                                                \end{table}

                                                \subsection{Contributions}
The main contributions are:
\begin{enumerate}
    \item A sensor-to-\texttt{world\_state} pipeline that combines a sensor-like perception stack with DriveLM-derived driving facts, converting language-grounded cues (e.g., sign readings) into typed fields with confidence.
    \item A three-server MCP tool layer (Rules, Weather, MCP-CAN) for compliance reasoning, traction-aware advisory, and vehicle-health context, with explicit rulepack versioning and auditable citations.
    \item A stateful orchestration design with parallel tool calls and an RSS/TTC-inspired safety arbiter that calibrates \emph{speak} versus \emph{act} decisions.
    \item A CARLA-based evaluation with VLM-only, VLM+RAG, and VLM+Tools baselines, plus robustness experiments (world-state corruption), arbiter sensitivity analysis, and CAN fault injection studies.
\end{enumerate}

Overall, (2) primarily addresses compliance, updateability, and traceability, (3) primarily addresses latency and robustness, (1) supports traceability/robustness at the perception--state boundary, and (4) provides empirical validation of robustness (and latency where measured).

\section{Related Work}
                                                                \subsection{Vision-Language Models and VLA Agents for Driving}
DriveGPT4 \cite{drivegpt4} demonstrates that a VLM can produce interpretable control predictions and explanations from multi-frame inputs.
DriveLM \cite{drivelm} frames driving with language as Graph Visual Question Answering (GVQA), explicitly modeling logical dependencies across perception, prediction, and planning tasks.
Beyond VQA-style formulations, recent work has explored vision-language planning and instruction-following agents for driving, including VLP \cite{vlp} and VLAAD \cite{vlaad}, as well as camera-only closed-loop VLA models such as SimLingo \cite{simlingo}.
Jiang \emph{et al.} present a systematic survey on current vision-language-action (VLA) models and their evaluation in autonomous driving \cite{vla_survey}.
These studies highlight the benefits of grounding and explanation, but many systems essentially merge multiple roles (perception, rule reasoning, planning) in a single inference.

DriveMCP adopts DriveLM as a perception-and-grounding front-end but decouples compliance reasoning, environment risk reasoning, and telemetry interpretation into external expert services with explicit interfaces.
In contrast to monolithic VLA policies, DriveMCP treats DriveLM outputs as typed evidence that is fused with tool responses and logged for audit, rather than as a direct control policy.

\subsection{Explainable Driving and Language Explanations}
Natural-language explanations for driving decisions have been studied as a way to improve transparency and user trust.
Kim \emph{et al.} introduced textual rationales and the BDD-X benchmark for self-driving explanations \cite{kim_eccv_2018}.
Another work explores retrieval-grounded explanations for improved factuality and generalization in driving assistants \cite{ragdriver}.
DriveMCP builds on this motivation by providing citations from jurisdiction-specific rulepacks and logging tool evidence to support post-drive audits.

\subsection{Retrieval-Augmented Generation (RAG) and External Knowledge}
Retrieval-augmented generation \cite{rag} augments parametric models with external memory by retrieving relevant documents at inference time.
For intelligent vehicles, external knowledge includes but is not limited to traffic codes, sign conventions, road authority directives (e.g., variable speed limits), and manufacturer guidance for service and repair of the vehicle.
DriveMCP uses RAG primarily for compliance and traceable citations: tool outputs include document identifiers and excerpts used for the final advisory.
For retrieval, hybrid pipelines commonly combine lexical methods such as BM25 \cite{bm25} with dense retrieval (e.g., DPR) \cite{dpr} to balance exact matches against semantic similarity.

\subsection{Agent Orchestration and Tool Protocols}
Tool protocols enable language models to call external services and devices.
MCP \cite{mcp} provides a standardized interface for tool discovery and structured call/response.
LangGraph-style orchestration \cite{langgraph} supports stateful, cyclic workflows with conditional branching and human-in-the-loop checkpoints.
More broadly, tool-use and modular reasoning patterns such as MRKL \cite{mrkl}, ReAct \cite{react}, and Toolformer \cite{toolformer} motivate separating symbolic/knowledge-intensive steps (e.g., compliance lookup) from perception and response generation.
DriveMCP uses a graph to coordinate perception updates, event extraction, tool routing, fusion, safety arbitration, and memory updates.

\subsection{Safety Guardrails and Standards Context}
Responsibility-Sensitive Safety (RSS) \cite{rss} provides an interpretable formalism for safe longitudinal and lateral interactions.
It has inspired open-source implementations~\cite{adrsslib} and is often used as a supervisory constraint layer.
Time-to-collision (TTC) has long been used as a risk indicator in traffic safety studies \cite{hayward_ttc}, and conservative braking-envelope models are standard in vehicle dynamics \cite{rajamani}.
In addition, functional safety and SOTIF considerations (e.g., ISO 26262 and ISO 21448) motivate conservative default behaviors, traceability, and explicit handling of uncertainties \cite{iso26262, iso21448}.
DriveMCP is not presented as a certified safety mechanism; rather, RSS/TTC checks provide transparent constraints for a research prototype.
In particular, the arbiter should be viewed as a transparent guardrail for experiments that constrain recommendations and optional interventions, not as a replacement for certified control and braking systems.
In a production setting, the same interfaces could be connected to certified safety monitors, and the full tool-trace (inputs, outputs, rulepack versions, and risk calculations) is intended to support post hoc analysis and SOTIF-style reasoning about uncertainty.

\section{System Overview}
\begin{figure*}[t]
\includegraphics[width=\textwidth]{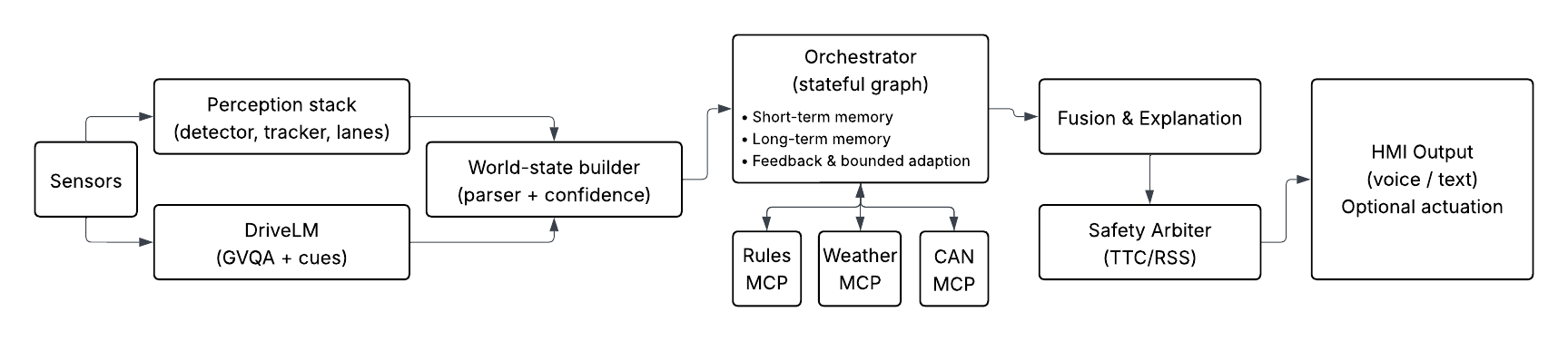}
\caption{DriveMCP architecture with sensor-like perception and DriveLM-derived fact parsing: camera and LiDAR feed a lightweight perception stack and DriveLM; a parser converts DriveLM cues into typed \texttt{world\_state} fields; an orchestration graph coordinates MCP servers (Rules, Weather, MCP-CAN); fused outputs are constrained by a safety arbiter; memory and feedback support adaptation and auditability.}
\label{fig:overview}
\end{figure*}

Fig.~\ref{fig:overview} articulates the overall architecture of DriveMCP.
A sensor interface provides synchronized camera, LiDAR, ego kinematics, and route context.
A lightweight perception stack produces actor tracks and lane cues, while DriveLM \cite{drivelm} processes the camera stream to produce GVQA answers and language-grounded cues.
Selected DriveLM outputs are parsed into typed fields and inserted into \texttt{world\_state} with confidence, so that the VLM is responsible for key driving facts (e.g., posted speed limits and jurisdiction cues) instead of receiving them as simulator ground truth.
A stateful orchestration graph (see Fig.~\ref{fig:overview}) then decides which MCP servers to call:
\begin{itemize}
    \item \textbf{Rules MCP server:} traffic code retrieval, sign interpretation, jurisdiction switching, and compliance advice.
    \item \textbf{Weather MCP server:} weather estimation (from simulation or API), traction assessment, and contextual speed and following-distance guidance.
    \item \textbf{MCP-CAN server:} decoded CAN signals, diagnostic trouble codes (DTCs), and health-aware context (e.g., ABS faults, low tire pressure).
\end{itemize}
Other MCP servers can be added by registering new tools behind the same MCP interface (e.g., map work zones, weather conditions, driver preference, or V2X alerts), without modifying the architecture.

Responses are fused into a single structured explanation and a recommended action.
A safety arbiter (see Fig.~\ref{fig:overview}) computes risk and applies guardrails to determine whether to \emph{speak} (advisory) or \emph{act} (intervention) and to enforce minimum safety constraints.
Finally, memory and feedback components store traces and enable bounded online adaptation.

Memory is accessed via an explicit read/write API.
Short-term memory caches recent \texttt{world\_state} snapshots and tool responses with timestamps and validity windows to reduce repeated calls and to enforce rate limits on messaging.
Long-term memory stores persistent artifacts such as \texttt{rulepack\_version} history, user preferences (e.g., advisory verbosity), and feedback labels.
The orchestrator queries memories at the start of each tick and commits a trace record after fusion and arbitration.

\section{DriveLM-Grounded Perception and World-State Construction}
\subsection{DriveLM and Graph VQA}
DriveLM \cite{drivelm} frames driving with language as a structured reasoning problem over graphs.
Instead of treating scene understanding as a flat captioning task, GVQA represents traffic participants, lanes, signs, and their relations in a graph with logical dependencies across tasks (e.g., perception $\rightarrow$ prediction $\rightarrow$ planning).
DriveLM provides datasets built on nuScenes \cite{nuscenes} and CARLA \cite{carla} and a baseline agent (DriveLM-Agent) that jointly performs GVQA and end-to-end driving.
These properties make DriveLM a suitable front-end for assistance systems that require both grounded understanding and interpretable intermediates.

\subsection{Sensor-like Perception Stack}
DriveMCP constructs \texttt{world\_state} from sensor-like observations that are closer to a production stack while remaining controlled and repeatable in CARLA.
We use a forward RGB camera and a LiDAR stream from CARLA, and we treat simulator ego kinematics as CAN-equivalent signals.
A lightweight perception stack detects and tracks nearby actors and provides lane context: (i) an off-the-shelf 2D detector on RGB for vehicles and pedestrians \cite{yolo}, (ii) LiDAR-based range refinement to estimate relative pose, (iii) a constant-velocity Kalman tracker for short-term tracking and drop-out handling \cite{kalman, sort}, and (iv) a lane and route interface from the CARLA map for lane assignment.
This stack intentionally avoids using simulator ground truth actor identities or sign metadata, so that downstream reasoning sees realistic imperfections.
The emphasis on explicit fusion and geometry is aligned with broader robotics practice; for example, prior work has used sensor fusion and perspective transformation to improve localization in embodied robots \cite{nadiri_ijira_localization_2025}.

\subsection{DriveLM-Derived Key Facts and Parsing}
Several compliance-critical fields are derived from DriveLM outputs rather than being injected directly.
For each frame, DriveLM is prompted with targeted GVQA questions about signs, speed limits, and jurisdiction cues (units, left-hand versus right-hand traffic, variable speed-limit gantries).
DriveLM produces language-grounded statements that are converted into typed fields using a constrained parser.
For example, we do not provide \texttt{speed\_limit\_kph=80} directly; instead the pipeline is:
\begin{equation}
\begin{aligned}
\text{image} \rightarrow \text{DriveLM} \rightarrow {} &
\text{sign reading} \rightarrow \text{parser} \\
& \rightarrow
\texttt{speed\_limit\_kph}=80.
\end{aligned}
\end{equation}
The parser performs numeric extraction, unit normalization, and schema validation, and it returns a confidence score.
Low-confidence parses trigger conservative fallbacks, such as issuing an advisory to slow down and routing a Rules query that explicitly asks for the applicable limit given the uncertain sign observation.

\subsection{World-State Representation and Uncertainty}
DriveMCP uses a typed \texttt{world\_state} with three groups of fields:
\begin{enumerate}
\item \textbf{Ego:} speed, acceleration, yaw rate, and lane assignment.
\item \textbf{Environment:} lanes, intersections, traffic lights, signs, dynamic speed limits, weather tags, and a jurisdiction tag.
\item \textbf{Actors:} a set of tracked objects with relative pose, class, and motion cues.
\end{enumerate}
Each field is represented as $(\text{value}, \text{confidence}, \text{provenance})$, where provenance records whether the field originated from the perception stack, a DriveLM parse, CAN/OBD telemetry, or the map, along with a timestamp.
Missing or low-confidence fields are left unset and are handled by conservative defaults in event extraction and the safety layer.
This design reduces the realism gap introduced by a fully ground-truth \texttt{world\_state} and makes failure modes attributable to specific upstream sources.
\subsection{Event Extraction for Tool Routing}
The orchestrator converts the current \texttt{world\_state} into discrete events that trigger tool calls.
Let $\phi(\cdot)$ be a feature function over \texttt{world\_state} and let $c(\cdot)$ denote associated confidences.
Event detectors output a set
\begin{equation}
E=\{e_i\mid g_i(\phi(\texttt{world\_state}),c(\texttt{world\_state}))>\eta_i\},
\end{equation}
where $g_i$ may be rule-based (e.g., a newly parsed sign with confidence above a threshold) or learned, and $\eta_i$ is a threshold.
Examples include \texttt{foreign\_sign}, \texttt{jurisdiction\_change}, \texttt{dynamic\_limit\_update}, \texttt{low\_friction}, and \texttt{vehicle\_fault}.
Confidence-aware triggers prevent brittle routing when perception is uncertain; for instance, if a speed-limit sign is detected but the DriveLM parse confidence is low, the system defaults to conservative advisory output and issues a Rules query that explicitly requests the applicable limit under uncertainty.
Event tags enable selective invocation: only relevant tools are called, reducing computation and driver distraction.

\section{MCP Tool Layer}
\subsection{Model Context Protocol (MCP)}
DriveMCP deploys specialized experts as MCP servers \cite{mcp}.
Each server exposes a tool catalog with JSON schemas, supports structured call/response, and can be instrumented for logging and safety policies (e.g., approval gates for high-impact actions).
This design reduces coupling between the orchestrator and expert implementations and enables incremental updates (e.g., adding a new region rule pack) without retraining the full system.
\subsection{Rules MCP Server with Retrieval}
The Rules server (see Fig.~\ref{fig:overview}) answers compliance questions such as:
\begin{itemize}
\item What does a sign mean in this jurisdiction and language?
\item Must the driver stop, yield, or proceed given right-of-way conventions?
\item What speed limit applies under a dynamic gantry or a ``when wet'' condition?
\end{itemize}

\begin{figure*}[t]
\includegraphics[width=\textwidth]{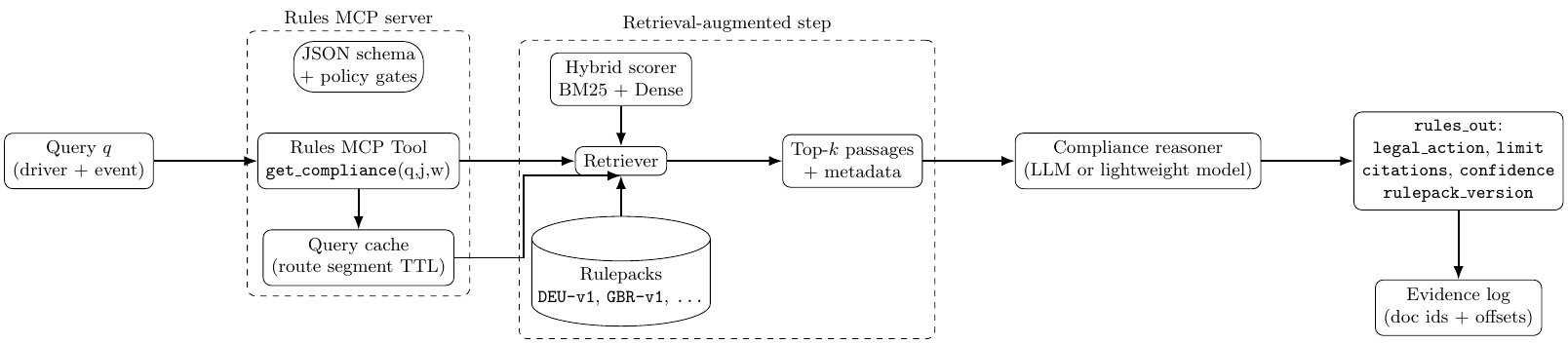}
\caption{Hybrid Retrieval-Augmented Generation (RAG) pipeline inside the Rules MCP server.}
\label{fig:rules_rag}
\end{figure*}
Fig.~\ref{fig:rules_rag} details the internal retrieval-augmented pipeline used by the Rules MCP server.
We implement the server as a retrieval-augmented pipeline \cite{rag}.
Given a driver query $q$, jurisdiction tag $j$, and world context $w$, the retrieval step returns documents $D=\texttt{Retrieve}(q,j,w)$.
A lightweight model then produces a structured response:
\begin{equation}
\begin{aligned}
\texttt{rules\_out}=\{&
\texttt{legal\_action}, \texttt{limit},\\
&\texttt{citations}, \texttt{confidence}
\}.
\end{aligned}
                                                                                                    \end{equation}

                                                                                                    Citations record document identifiers and offsets for auditability.
                                                                                                    In multilingual cases, the server includes translated sign meaning and clarifies unit conversions.

                                                                                                    \subsubsection{Hybrid Retrieval and Corpus Management}
                                                                                                    Traffic regulation queries often require both semantic similarity (e.g., ``right-of-way at a roundabout'') and exact lexical matches (e.g., a sign code or a statutory phrase).
                                                                                                    We therefore support a hybrid retrieval score that combines sparse and dense signals:
                                                                                                    \begin{equation}
                                                                                                    S(q,d)=(1-\lambda)\,\text{BM25}(q,d)+\lambda\,\cos(\phi(q),\phi(d)),
                                                                                                    \end{equation}
                                                                                                    where $\phi(\cdot)$ is an embedding model and $\lambda\in[0,1]$ balances the two terms.
                                                                                                    In practice, $\lambda$ is tuned per corpus (statutes vs.\ sign tables) and cached for reproducibility.
                                                                                                    The Rules server also tracks a \texttt{rulepack\_version} so that advisories can be audited against the specific code snapshot used at inference time.
                                                                                                    In our implementation, sparse retrieval uses BM25-style indexing \cite{bm25}, while dense retrieval can use DPR-like embeddings for semantic matching \cite{dpr}.

                                                                                                    \subsubsection{Rulepacks, Coverage, and Versioning}
                                                                                                    The Rules server retrieves from jurisdiction-scoped \emph{rulepacks}.
                                                                                                    A rulepack is a versioned bundle of (i) traffic-code excerpts, (ii) sign tables and conventions, and (iii) unit metadata (km/h versus mph, left-hand versus right-hand traffic), stored as JSON plus the original source documents.
                                                                                                    In practice, sign conventions can be grounded in widely used standards and official guidance, including the Vienna Convention on Road Signs and Signals \cite{vienna_convention}, the U.S.\ MUTCD \cite{mutcd}, and the U.K.\ TSRGD \cite{tsrgd}.
                                                                                                    In our experiments we use three rulepacks: \texttt{DEU-v1} (km/h, right-hand traffic), \texttt{GBR-v1} (mph, left-hand traffic), and \texttt{DEU-v1-VSL} for variable speed-limit gantries and ``when wet'' clauses.
                                                                                                    Each rulepack is versioned with semantic versions and a content hash.
                                                                                                    All Rules outputs include \texttt{rulepack\_version} and document identifiers to support audit trails and regression testing.

                                                                                                    When multiple passages yield conflicting advice, the Rules server applies an explicit priority order (statutory code, official sign table, then secondary guidance) and defaults to the most conservative limit when ambiguity remains.
                                                                                                    Rulepack staleness is handled via \texttt{effective\_date} metadata: if the active vehicle configuration indicates a newer rulepack than the retrieved passage, the server lowers confidence and the orchestrator defaults to conservative advisory output.
                                                                                                    We provide a small synthetic rulepack and accompanying unit tests in the supplemental material to support reproducibility and auditability.

                                                                                                    \subsubsection{Jurisdiction Switching and Unit Normalization}
                                                                                                    Cross-border scenarios require switching between rule sets and normalizing units (e.g., mph vs.\ km/h).
                                                                                                    DriveMCP represents jurisdiction as an explicit field in \texttt{world\_state} and treats a jurisdiction change as a first-class event.
                                                                                                    The Rules server returns both the legal limit and a normalized limit in SI units, enabling consistent downstream safety computation.
                                                                                                    When signage conventions differ (e.g., sign shapes or languages), the server includes a short explanatory note to prevent driver confusion.

                                                                                                    \subsubsection{Caching and Timeouts}
                                                                                                    To meet real-time constraints, DriveMCP caches frequent rule queries within a route segment (e.g., repeated speed-limit reminders) and uses per-tool timeouts.
                                                                                                    If retrieval fails or a tool times out, the orchestrator issues a conservative advisory based on available context and logs the degradation for later analysis.

                                                                                                    \subsection{Weather MCP Server}
                                                                                                    The Weather server provides precipitation intensity, visibility risk, and an estimated friction range $\mu_{\text{est}}$.
                                                                                                    In simulation, these values are derived from CARLA weather parameters; in a vehicle, they can combine onboard sensors and external APIs.
                                                                                                    The server outputs:
                                                                                                    \begin{equation}
                                                                                                    \begin{aligned}
                                                                                                    \texttt{wx\_out}=\{&
                                                                                                    \mu_{\text{est}}, \texttt{risk\_level},\\
                                                                                                    &\texttt{speed\_factor}, \texttt{confidence}
                                                                                                    \}.
                                                                                                    \end{aligned}
                                                                                                    \end{equation}

                                                                                                    The \texttt{speed\_factor} scales legal limits into traction-aware advisories; it can also adjust safe following-distance parameters in the safety layer.
 \subsection{MCP-CAN Server}
                                                                                                    Vehicle telemetry provides critical context not visible in camera images, including actuator limits, diagnostic faults, and driver inputs.
                                                                                                    We integrate an MCP-CAN server \cite{mcp_can} that bridges CAN signals and OBD-II diagnostics into MCP tools, built on top of the CAN protocol family \cite{iso11898} and standard OBD-II diagnostic services (e.g., SAE J1979) \cite{sae_j1979}.
                                                                                                    The server supports decoding CAN frames via DBC definitions, querying selected signals, and reading diagnostic trouble codes (DTCs).
                                                                                                    For safety, DriveMCP treats CAN tools as read-only by default; any actuation-capable command (if enabled) is gated by explicit confirmation and the safety arbiter.

                                                                                                    \subsubsection{Signal Selection and Health Indicators}
                                                                                                    Modern vehicles expose hundreds of signals; a driver-assistance layer should request only a minimal subset to reduce bandwidth and privacy risk.
                                                                                                    DriveMCP uses a curated signal list grouped by function: longitudinal control (speed, acceleration, brake pressure), lateral status (steering angle, stability control), and health indicators (DTC flags, warning lamps, tire pressure where available).
                                                                                                    The orchestrator requests signals only when a relevant event is present (e.g., \texttt{vehicle\_fault}) or when computing the health score $h$ at a low rate.

                                                                                                    \subsubsection{Diagnostics-to-Explanation Mapping}
                                                                                                    Diagnostic trouble codes are often cryptic.
                                                                                                    The MCP-CAN server provides a translation layer from codes to user-facing summaries (e.g., ``powertrain fault detected'') and recommended next steps (e.g., ``reduce speed; seek service soon'').
                                                                                                    DriveMCP treats these as advisory content only; it does not attempt to perform repair actions.

                                                                                                    \subsection{Security and Policy Considerations}
                                                                                                    Tool modularity increases the attack surface; therefore, DriveMCP assigns each tool a capability level.
                                                                                                    Context-only tools (Rules, Weather, CAN-read) are always enabled.
                                                                                                    Potentially dangerous tools (e.g., those that could affect actuation) are disabled by default and require out-of-band authorization.
                                                                                                    All tool calls are logged with timestamps and hashed payloads to support audit trails.

                                                                                                    \section{Orchestration, Fusion, and Safety Arbitration}
                                                                                                    \subsection{Stateful Graph Orchestration}
                                                                                                    DriveMCP orchestrates tool use via a stateful graph \cite{langgraph}.
                                                                                                    Each node reads/writes a subset of state $\mathcal{S}$:
                                                                                                    \begin{equation}
                                                                                                    \begin{aligned}
                                                                                                    \mathcal{S}=\{&
                                                                                                    \texttt{world\_state},\texttt{events},\texttt{tool\_outs},\texttt{fused},\\
                                                                                                    &\texttt{safety},\texttt{decision},\texttt{memory}
                                                                                                    \}.
                                                                                                    \end{aligned}
                                                                                                    \end{equation}

                                                                                                    Edges implement conditional routing based on event types and confidence.
                                                                                                    When multiple factors are present (e.g., speed limit sign + rain + DTC), the orchestrator invokes Rules, Weather, and CAN tools in parallel and waits for completion or timeouts.
                                                                                                    This design improves latency and avoids irrelevant tool calls.
\textbf{Memory interfaces:} The orchestrator communicates with memory modules through (i) a short-term cache used for deduplication, rate limiting, and reuse of recent tool evidence, and (ii) a long-term store used for preference, feedback, and audit trails.
Concretely, nodes can read cached tool outputs before issuing a call, and every tick writes an append-only trace that records inputs, tool calls, responses, and the final arbitration decision.

\textbf{Temporal aspects and time-sensitive queries:} Each orchestration tick runs under a fixed end-to-end budget that reflects the dynamics of driving assistance.
Tool calls are issued in parallel with per-tool timeouts.
If a query is time sensitive (e.g., variable speed limits, sudden weather changes, or a rapidly closing lead vehicle), late tool responses are ignored for the current tick and can be incorporated on the next tick.
\texttt{world\_state} fields include timestamps and validity windows so that stale facts are either refreshed (re-query DriveLM or sensors) or treated with reduced confidence, and the arbiter defaults to the most conservative safe action when uncertainty is high.

                                                                                                    Algorithm~\ref{alg:tick} sketches one orchestration tick.

                                                                                                    \begin{algorithm}[t]
                                                                                                    \caption{DriveMCP orchestration tick (simplified).}
                                                                                                    \label{alg:tick}
                                                                                                    \begin{algorithmic}[1]
                                                                                                    \STATE Input: sensor snapshot, driver query
                                                                                                    \STATE Update \texttt{world\_state} using simulator/DriveLM adapter
                                                                                                    \STATE $E \leftarrow$ extract events from \texttt{world\_state}
                                                                                                    \STATE Route: select MCP tools based on $E$
                                                                                                    \STATE Parallel call: Rules, Weather, CAN tools (with timeouts)
                                                                                                    \STATE Fuse tool outputs into \texttt{fused} response + confidence
                                                                                                    \STATE Compute safety metrics (TTC, RSS violation) from \texttt{world\_state}
                                                                                                    \STATE Decide SPEAK vs ACT using guardrails and thresholds
                                                                                                    \STATE Log trace; update memory; apply bounded feedback updates
                                                                                                    \STATE Output: advisory (and optional bounded actuation)
                                                                                                    \end{algorithmic}
                                                                                                    \end{algorithm}

                                                                                                    \subsection{Fusion and Conflict Handling}
Tool outputs are fused into a single explanation and action.
Legal compliance has highest priority; safety and traction refine the recommended target (e.g., advising below the legal limit).
CAN-derived health constraints can further reduce recommended speed or increase headway.
When there is a contradiction (e.g., legal limit vs.\ traction), DriveMCP communicates both: the \emph{legal requirement} and a \emph{recommended safe target}, with an explanation of why.

\subsection{Safety Metrics and RSS-Inspired Guardrails}
                                                                                                    We compute safety metrics from \texttt{world\_state}: time-to-collision (TTC), headway, and an RSS-style violation flag.
                                                                                                    For a lead vehicle at relative distance $d$ and closing speed $\Delta v$:
                                                                                                    \begin{equation}
                                                                                                    \text{TTC}=\frac{d}{\max(\epsilon,\Delta v)}.
                                                                                                    \end{equation}
                                                                                                    A simplified RSS-inspired check flags unsafe gaps:
                                                                                                    \begin{equation}
                                                                                                    \texttt{rss\_viol}=\mathbb{1}[d<d_{\min}(v,\mu_{\text{est}})].
                                                                                                    \end{equation}
                                                                                                    Here $d_{\min}$ accounts for reaction time and braking limits; in the prototype we use a conservative parametric form and treat $\mu_{\text{est}}$ as uncertain under rain.
                                                                                                    TTC-based criteria have a long history in traffic safety \cite{hayward_ttc}, and we use them here only as an interpretable risk surrogate for a research prototype.

                                                                                                    \subsection{Health-Aware Risk Shaping from CAN}
                                                                                                    CAN context affects both risk and messaging.
                                                                                                    We derive a health score $h\in[0,1]$ from CAN/OBD signals:
                                                                                                    \begin{equation}
                                                                                                    h = \sigma\left(\sum_k w_k \cdot \psi_k(\texttt{can\_signals})\right),
                                                                                                    \end{equation}
                                                                                                    where $\psi_k$ are normalized indicators (e.g., ABS fault present, brake temperature high, low tire pressure) and $\sigma$ is a squashing function.
                                                                                                    The risk score is:
                                                                                                    \begin{equation}
                                                                                                    r=\alpha \max\left(0,\frac{\tau-\text{TTC}}{\tau}\right)+\beta\cdot \texttt{rss\_viol}+\gamma\cdot h.
                                                                                                    \end{equation}
                                                                                                    This mechanism makes the system more conservative under degraded health, without relying on implicit model behavior.

                                                                                                    \subsection{Speak vs.\ Act Arbitration}
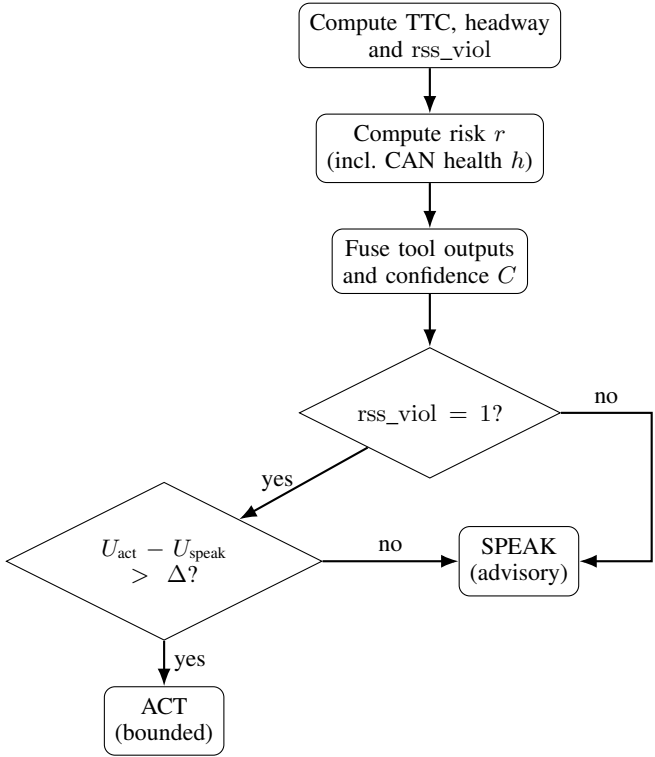
\begin{figure}[t]
                                                                                                    \centering
                                                                                                    \begin{tikzpicture}[
                                                                                                      node distance=6mm,
                                                                                                        font=\small,
                                                                                                          proc/.style={draw, rounded corners, align=center, inner sep=4pt},
                                                                                                            dec/.style={draw, diamond, align=center, inner sep=1pt, text width=28mm, aspect=2},
                                                                                                              arr/.style={-Latex, thick}
                                                                                                              ]
                                                                                                              % Nodes
                                                                                                              \node[proc] (s0) {Compute TTC, headway\\and $\mathrm{rss\_viol}$};
                                                                                                              \node[proc, below=of s0] (s1) {Compute risk $r$\\(incl.\ CAN health $h$)};
                                                                                                              \node[proc, below=of s1] (s2) {Fuse tool outputs\\and confidence $C$};

                                                                                                              \node[dec, below=of s2, yshift=-1mm] (d0) {$\mathrm{rss\_viol}=1$?};

                                                                                                              \node[dec, below left=10mm and 16mm of d0] (d1)
                                                                                                                {$U_{\text{act}}-U_{\text{speak}}$\\$>\Delta$?};

                                                                                                                \node[proc, right=18mm of d1] (speak) {SPEAK\\(advisory)};
                                                                                                                \node[proc, below=of d1] (act) {ACT\\(bounded)};

                                                                                                                % Main chain
                                                                                                                \draw[arr] (s0) -- (s1);
                                                                                                                \draw[arr] (s1) -- (s2);
                                                                                                                \draw[arr] (s2) -- (d0);

                                                                                                                % Decisions
                                                                                                                % d0: yes -> utility check, no -> speak
                                                                                                                \draw[arr] (d0) -- node[left] {yes} (d1);

                                                                                                                % Force a clean horizontal "no" branch first, then down to SPEAK
                                                                                                                \draw[arr] (d0.east) -- ++(12mm,0)
                                                                                                                  node[midway,above] {no}
                                                                                                                    |- (speak.east);

                                                                                                                    % d1: yes -> act, no -> speak
                                                                                                                    \draw[arr] (d1) -- node[right] {yes} (act);
                                                                                                                    \draw[arr] (d1.east) -- node[above] {no} (speak.west);

                                                                                                                    \end{tikzpicture}%

                                                                                                                    \caption{Safety arbitration: actuation is permitted only when an RSS-style safety condition is violated and the utility margin exceeds $\Delta$; otherwise the system defaults to advisory output.}
                                                                                                                    \label{fig:arbit}
                                                                                                                    \end{figure}

                                                                                                    DriveMCP treats interventions as higher-cost actions than advisories.
                                                                                                    We define utilities:
                                                                                                    \begin{align}
                                                                                                    U_{\text{act}} &= r - C_{\text{act}} - C_{\text{err}},\\
                                                                                                    U_{\text{speak}} &= B_{\text{adv}}\cdot r - C_{\text{fat}}.
                                                                                                    \end{align}
                                                                                                    A decision margin $\Delta$ is used:
                                                                                                    \begin{equation}
                                                                                                    \texttt{decision}=
                                                                                                    \begin{cases}
                                                                                                    \texttt{ACT}, & (U_{\text{act}}-U_{\text{speak}})>\Delta \wedge \texttt{rss\_viol}=1,\\
                                                                                                    \texttt{SPEAK}, & \text{otherwise.}
                                                                                                    \end{cases}
                                                                                                    \end{equation}
                                                                                                    This enforces that actuation is only permitted when an interpretable safety condition is violated.
                                                                                                    Fig.~\ref{fig:arbit} visualizes the guardrail logic.

                                                                                                                    \subsection{Arbiter Parameterization and Calibration}
                                                                                                                    DriveMCP explicitly parameterizes the safety arbiter so that behavior can be audited and tuned without retraining perception or language components.
                                                                                                                    Table~\ref{tab:arb_params} lists the constants used in all experiments.
                                                                                                                    We compute the RSS-inspired minimum gap using a conservative longitudinal model:
                                                                                                                    \begin{equation}
                                                                                                                    d_{\min}(v,\mu_{\text{est}})= v\,t_r + \frac{v^2}{2\,a_{\text{brake}}(\mu_{\text{est}})} + d_0,\quad a_{\text{brake}}(\mu_{\text{est}})=\mu_{\text{est}}\,g,
                                                                                                                    \end{equation}
where $t_r$ is reaction time, $d_0$ is a standstill buffer, and we use the lower bound of $\mu_{\text{est}}$ under rain for conservatism.
We set $t_r=0.7$~s as a conservative human-reaction-time assumption commonly used in ADAS-style safety models.
                                                                                                                    This braking model is intentionally simplified; more detailed approaches are standard in vehicle dynamics \cite{rajamani}.

                                                                                                                    All arbiter weights and margins ($\alpha,\beta,\gamma,\tau,\Delta$ and utility constants) are tuned offline on a held-out calibration set that mixes all three scenario families.
                                                                                                                    During evaluation, all parameters are fixed across scenarios; we do not tune per scenario.

                                                                                                                    \begin{table}[t]
                                                                                                                    \centering
                                                                                                                    \caption{Safety arbiter constants used in all experiments.}
                                                                                                                    \label{tab:arb_params}
                                                                                                                    \setlength{\tabcolsep}{4pt}
                                                                                                                    \begin{tabular}{p{0.20\columnwidth}p{0.18\columnwidth}p{0.52\columnwidth}}
                                                                                                                    \toprule
                                                                                                                    Parameter & Value & Meaning \\
                                                                                                                    \midrule
                                                                                                                    $\epsilon$ & 0.1 & TTC denominator floor \\
                                                                                                                    $t_r$ & 0.7 s & reaction time in $d_{\min}$ \\
                                                                                                                    $d_0$ & 2.0 m & standstill buffer in $d_{\min}$ \\
                                                                                                                    $\tau$ & 2.5 s & TTC reference in risk score $r$ \\
                                                                                                                    $\alpha,\beta,\gamma$ & 0.6, 0.3, 0.1 & weights for TTC term, RSS flag, CAN health \\
                                                                                                                    $\Delta$ & 0.2 & actuation margin in speak versus act \\
                                                                                                                    $C_{\text{act}}$ & 0.4 & cost of intervention \\
                                                                                                                    $C_{\text{err}}$ & 0.2 & penalty for erroneous intervention \\
                                                                                                                    $B_{\text{adv}}$ & 0.6 & advisory benefit factor \\
                                                                                                                    $C_{\text{fat}}$ & 0.05 & driver fatigue cost \\
                                                                                                                    \bottomrule
                                                                                                                    \end{tabular}
                                                                                                                    \end{table}

                                                                                                                    \subsection{Feedback and Bounded Adaptation}
                                                                                                                    DriveMCP logs tool calls, fused outputs, safety metrics, and decisions.
                                                                                                                    Feedback events (explicit driver ratings, overrides, or supervisor labels such as \texttt{late\_warning} or \texttt{false\_positive}) update parameters within bounds:
                                                                                                                    \begin{align}
                                                                                                                    \Delta &\leftarrow \left[\Delta-k_{\text{late}}\,n_{\text{late}}\right]_{\Delta_{\min}}^{\Delta_{\max}},\\
                                                                                                                    C_{\min} &\leftarrow \left[C_{\min}+k_{\text{fp}}\,n_{\text{fp}}\right]_{C_{\min}}^{C_{\max}}.
                                                                                                                    \end{align}
                                                                                                                    These monotone updates reduce late warnings while controlling false positives and avoiding oscillations.
                                                                                                                    Offline calibration periodically updates default parameters from aggregated traces.

                                                                                                                    \section{Implementation}
                                                                                                                    \subsection{Process Architecture}
                                                                                                                    DriveMCP runs as multiple processes communicating via RPC:
                                                                                                                    (1) orchestrator, (2) retrieval service, (3) Rules server, (4) Weather server, (5) MCP-CAN server, and (6) logging/memory service.
                                                                                                                    This separation supports isolation and independent updates.

                                                                                                                    \subsection{Latency Optimizations}
                                                                                                                    To maintain responsiveness, DriveMCP uses:
                                                                                                                    \begin{itemize}
                                                                                                                        \item \textbf{Selective invocation:} tools are called only when events require them.
                                                                                                                            \item \textbf{Parallel MCP calls:} multi-factor situations are handled concurrently.
                                                                                                                                \item \textbf{Caching:} repeated queries (e.g., the same stop sign on a short segment) reuse retrieved passages.
                                                                                                                                    \item \textbf{Timeout and fallback:} if a tool times out, the system degrades gracefully to SPEAK-only conservative advice.
                                                                                                                                    \end{itemize}

                                                                                                                                    \section{Experimental Setup}
                                                                                                                                    \subsection{Simulation Environment and Scenarios}
                                                                                                                                    Experiments are conducted in CARLA 0.9.8 \cite{carla} across multiple towns under day/night lighting and controllable weather.
                                                                                                                                    We evaluate three scenario families that stress compliance and adaptation:
                                                                                                \begin{figure}[t]
\centering
\includegraphics[width=\linewidth]{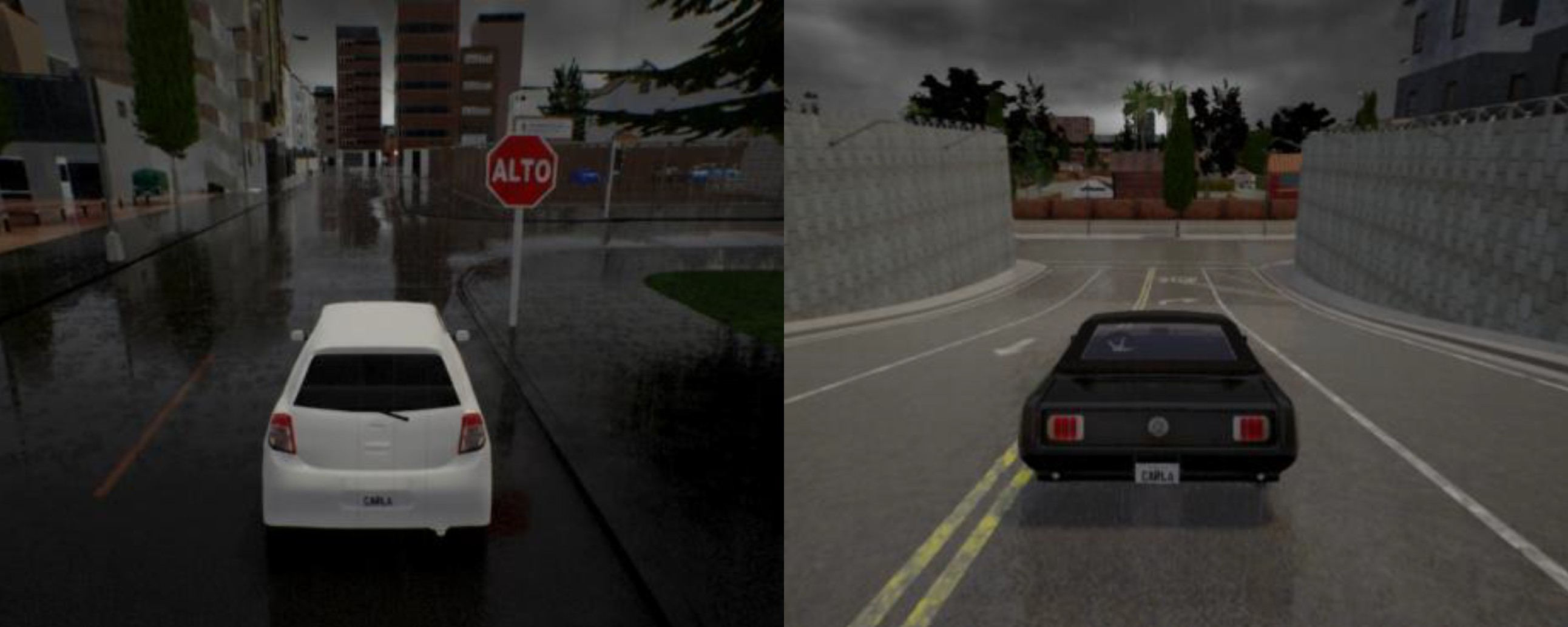}
\caption{Scenario visualizations. Left: urban rain multilingual, Right: dynamic variable speed limit.}
\label{fig:scenario_visualizations}
\end{figure}
                                                                                                
                                                                                                       \begin{enumerate}          \item \textbf{URB-RAIN-MULTILINGUAL:} urban routes with heavy rain and foreign-language signage; the assistant must interpret signs and recommend safe speed and headway.
                                                                                                                                            \item \textbf{CROSS-BORDER:} routes that cross jurisdictions with changes in signage conventions, right/left-hand traffic, and units (km/h vs.\ mph); the assistant must switch rule sets and explanations.
                                                                                                                                                \item \textbf{DYNAMIC-VSL:} corridors with variable speed limit gantries and weather-conditioned limits; the assistant must adopt new limits quickly while accounting for traction.
                                                                                                                                                \end{enumerate}
                                                                                                                                                For comparisons, perception inputs are constructed from the sensor-like stack in Section~IV, including DriveLM-parsed speed limits and jurisdiction cues.
                                                                                                                                                Unless otherwise stated, all methods operate on the same sensor snapshots and the same rulepack configuration.
                                                                                                                                                To quantify remaining realism gaps, we also evaluate robustness under controlled \texttt{world\_state} corruption (missed signs, misread limits, and actor dropouts) and report an oracle variant that uses simulator metadata only for analysis.

                                                                                                                                                \subsection{Rulepacks and Jurisdictions}
                                                                                                                                                We evaluate DriveMCP under explicit jurisdiction configurations to support auditability and reproducibility.
                                                                                                                                                Table~\ref{tab:rulepacks} lists the rulepacks used in our scenarios and their unit and traffic-side conventions.

                                                                                                                                                \begin{table}[t]
                                                                                                                                                \centering
                                                                                                                                                \caption{Rulepacks and jurisdiction configurations used in experiments.}
                                                                                                                                                \label{tab:rulepacks}
                                                                                                                                                \setlength{\tabcolsep}{4pt}
                                                                                                                                                \begin{tabular}{p{0.20\columnwidth}p{0.15\columnwidth}p{0.18\columnwidth}p{0.32\columnwidth}}
                                                                                                                                                \toprule
                                                                                                                                                Rulepack & Units & Traffic side & Used in scenarios \\
                                                                                                                                                \midrule
                                                                                                                                                \texttt{DEU-v1} & km/h & right-hand & URB-RAIN, DYNAMIC-VSL (static limits) \\
                                                                                                                                                \texttt{DEU-v1-VSL} & km/h & right-hand & DYNAMIC-VSL (gantries, when-wet) \\
                                                                                                                                                \texttt{GBR-v1} & mph & left-hand & CROSS-BORDER (post-switch segment) \\
                                                                                                                                                \bottomrule
                                                                                                                                                \end{tabular}
                                                                                                                                                \end{table}

                                                                                                                                                \subsection{Models, Context Inputs, and Prompts}
                                                                                                                                                DriveMCP uses DriveLM \cite{drivelm} as the vision-language front end for GVQA and for extracting language-grounded driving facts (sign readings, jurisdiction cues, variable limit updates).
                                                                                                                                                For all VLM-based baselines we use the same DriveLM checkpoint to isolate the effects of retrieval, tool use, and arbitration.

                                                                                                                                                All methods receive the same synchronized sensor snapshot (RGB image, LiDAR, ego kinematics, and route context) and the same actor list from the perception stack.
                                                                                                                                                For VLM-only baselines, the structured \texttt{world\_state} is converted to a compact textual summary that is appended to the prompt.
                                                                                                                                                For tool-based methods, \texttt{world\_state} is passed as structured JSON to tool calls.

                                                                                                                                                Prompt structure is fixed across methods: (i) a short system instruction specifying that the assistant must be conservative, (ii) a context block containing ego state and a perception summary, (iii) the driver query, and (iv) optional retrieved passages or tool outputs.
                                                                                                                                                VLM-Direct and VLM-Direct+RAG are restricted to advisory-only output via prompting and do not emit actuation commands.
                                                                                                                                                VLM-Tools-NoArbiter and DriveMCP can produce bounded actuation suggestions, but only DriveMCP filters them through the RSS/TTC arbiter.

                                                                                                                                                \subsection{World-State Corruption Protocol}
                                                                                                                                                To quantify robustness and to close the realism gap introduced by clean simulator state, we inject controlled corruption into selected \texttt{world\_state} fields derived from perception.
                                                                                                                                                We consider three corruption types: (i) missed signs (drop the speed-limit sign observation), (ii) misread limits (perturb the parsed limit value and units), and (iii) actor dropouts (remove a tracked actor for a time window).
                                                                                                                                                We measure how corruption degrades (a) event extraction and tool routing (event F1 and correct-tool rate), and (b) downstream safety and compliance metrics (infractions, overspeed, hazard response, and TTC at intervention).

                                                                                                                                                \subsection{Baselines}
                                                                                                                                                We compare DriveMCP against three VLM baselines that isolate the impact of retrieval, tools, and arbitration:
                                                                                                                                                \begin{itemize}
                                                                                                                                                    \item \textbf{VLM-Direct:} a single DriveLM inference that consumes the current image and a textual perception summary and outputs an advisory response.
                                                                                                                                                        \item \textbf{VLM-Direct+RAG:} the same as VLM-Direct, but with top-$k$ retrieved traffic-code passages and sign-table snippets concatenated to the prompt.
                                                                                                                                                            \item \textbf{VLM-Tools-NoArbiter:} a tool-augmented variant that uses the same MCP tool layer (Rules, Weather, MCP-CAN) and the same event-driven routing as DriveMCP, but bypasses the safety arbiter and returns the fused recommendation directly.
                                                                                                                                                                \item \textbf{DriveMCP (ours):} full modular system with Rules, Weather, and CAN context, plus safety arbitration and memory and feedback.
                                                                                                                                                                \end{itemize}

                                                                                                                                                                \begin{table}[t]
                                                                                                                                                                \centering
                                                                                                                                                                \caption{Baseline configurations. All methods receive the same synchronized sensor snapshot.}
                                                                                                                                                                \label{tab:baseline_cfg}
                                                                                                                                                                \setlength{\tabcolsep}{4pt}
                                                                                                                                                                \begin{tabular}{p{0.20\columnwidth}p{0.14\columnwidth}p{0.10\columnwidth}p{0.16\columnwidth}p{0.22\columnwidth}}
                                                                                                                                                                \toprule
                                                                                                                                                                Method & Retrieval & Tools & Safety arbiter & \texttt{world\_state} form \\
                                                                                                                                                                \midrule
                                                                                                                                                                VLM-Direct & no & no & no & text summary \\
                                                                                                                                                                VLM-Direct+RAG & yes & no & no & text summary \\
                                                                                                                                                                VLM-Tools-NoArbiter & yes & yes & no & structured JSON \\
                                                                                                                                                                DriveMCP (ours) & yes & yes & yes & structured JSON \\
                                                                                                                                                                \bottomrule
                                                                                                                                                                \end{tabular}
                                                                                                                                                                \end{table}

                                                                                                                                                                \subsection{Metrics}
                                                                                                                                                                We report:
                                                                                                                                                                \begin{itemize}
                                                                                                                                                                    \item \textbf{Infractions per 10 km:} speeding, failure to stop, illegal turns, incorrect yielding.
                                                                                                                                                                        \item \textbf{Overspeed (\%):} average percentage above the inferred posted or dynamic limit.
                                                                                                                                                                            \item \textbf{Hazard response time (s):} delay from hazard onset to first effective warning or intervention.
                                                                                                                                                                                \item \textbf{Advisory latency (s):} end-to-end time from event detection to advisory generation.
                                                                                                                                                                                    \item \textbf{TTC at first escalation (s):} TTC at the moment the system first escalates; higher indicates earlier escalation.
                                                                                                                                                                                        \item \textbf{Event extraction F1 (\%):} F1 score for detecting key discrete events.
                                                                                                                                                                                            \item \textbf{Correct-tool rate (\%):} fraction of timesteps where the system invokes the correct subset of tools.
                                                                                                                                                                                                \item \textbf{Intervention false-positive rate (\%):} fraction of interventions that occur without an RSS violation.
                                                                                                                                                                                                    \item \textbf{Qualitative ratings:} explanation clarity and perceived helpfulness collected via debriefing questionnaires.
                                                                                                                                                                                                    \end{itemize}

                                                                                                                                                                                                    \section{Results}
                                                                                                                                                                                                    Table~\ref{tab:results} summarizes results for the three scenario families.
                                                                                                                                                                                                    VLM-Direct benefits from language grounding but lacks explicit rule retrieval, tool access, and arbitration, which leads to more infractions under jurisdiction changes and dynamic limits.
                                                                                                                                                                                                    Adding retrieval (VLM-Direct+RAG) improves compliance, especially in CROSS-BORDER and DYNAMIC-VSL, but increases latency due to longer context and does not consistently improve hazard response.
                                                                                                                                                                                                    Tool augmentation (VLM-Tools-NoArbiter) improves both compliance and hazard response, but it can trigger unnecessary escalations without an explicit safety filter.
                                                                                                                                                                                                    DriveMCP achieves the lowest infraction rate and overspeed across scenarios while maintaining sub-second advisory latency through selective routing and parallel MCP calls.

                                                                                                                                                                                                    \begin{table*}[t]
                                                                                                                                                                                                    \centering
                                                                                                                                                                                                    \caption{Performance in CARLA scenarios (mean $\pm$ std.\ over 10 runs). Lower is better for infractions, overspeed, hazard response, and latency; higher is better for TTC at first escalation.}
                                                                                                                                                                                                    \label{tab:results}
                                                                                                                                                                                                    \footnotesize
                                                                                                                                                                                                    \setlength{\tabcolsep}{4pt}
                                                                                                                                                                                                    \begin{tabular*}{\textwidth}{@{\extracolsep{\fill}}lccccc}
                                                                                                                                                                                                    \toprule
                                                                                                                                                                                                    Method &
                                                                                                                                                                                                    \makecell{Infractions/10 km\\$\downarrow$} &
                                                                                                                                                                                                    \makecell{Overspeed (\%)\\$\downarrow$} &
                                                                                                                                                                                                    \makecell{Hazard response (s)\\$\downarrow$} &
                                                                                                                                                                                                    \makecell{Advisory latency (s)\\$\downarrow$} &
                                                                                                                                                                                                    \makecell{TTC at escalation (s)\\$\uparrow$} \\
                                                                                                                                                                                                    \midrule

                                                                                                                                                                                                    \multicolumn{6}{l}{\textbf{URB-RAIN-MULTILINGUAL}}\\
                                                                                                                                                                                                    VLM-Direct & $1.1\pm0.3$ & 2.1 & 1.10 & 0.68 & 1.5 \\
                                                                                                                                                                                                    VLM-Direct+RAG & $0.9\pm0.3$ & 1.8 & 1.05 & 0.78 & 1.6 \\
                                                                                                                                                                                                    VLM-Tools-NoArbiter & $0.4\pm0.2$ & 1.0 & 0.70 & 0.25 & 2.2 \\
                                                                                                                                                                                                    DriveMCP (ours) & $0.3\pm0.2$ & 0.9 & 0.73 & 0.29 & 2.0 \\
                                                                                                                                                                                                    \midrule

                                                                                                                                                                                                    \multicolumn{6}{l}{\textbf{CROSS-BORDER}}\\
                                                                                                                                                                                                    VLM-Direct & $0.9\pm0.4$ & 1.6 & 1.02 & 0.61 & 1.6 \\
                                                                                                                                                                                                    VLM-Direct+RAG & $0.5\pm0.2$ & 1.2 & 0.95 & 0.71 & 1.7 \\
                                                                                                                                                                                                    VLM-Tools-NoArbiter & $0.3\pm0.1$ & 0.8 & 0.70 & 0.27 & 2.3 \\
                                                                                                                                                                                                    DriveMCP (ours) & $0.2\pm0.1$ & 0.7 & 0.72 & 0.31 & 2.1 \\
                                                                                                                                                                                                    \midrule

                                                                                                                                                                                                    \multicolumn{6}{l}{\textbf{DYNAMIC-VSL}}\\
                                                                                                                                                                                                    VLM-Direct & $0.9\pm0.3$ & 1.8 & 1.08 & 0.66 & 1.6 \\
                                                                                                                                                                                                    VLM-Direct+RAG & $0.5\pm0.2$ & 1.4 & 1.00 & 0.74 & 1.7 \\
                                                                                                                                                                                                    VLM-Tools-NoArbiter & $0.3\pm0.2$ & 0.7 & 0.67 & 0.26 & 2.2 \\
                                                                                                                                                                                                    DriveMCP (ours) & $0.2\pm0.2$ & 0.6 & 0.69 & 0.30 & 2.0 \\

                                                                                                                                                                                                    \bottomrule
                                                                                                                                                                                                    \end{tabular*}
                                                                                                                                                                                                    \end{table*}

                                                                                                                                                                                                    \begin{figure*}[t]
                                                                                                                                                                                                    \centering
                                                                                                                                                                                                    \includegraphics[width=\textwidth]{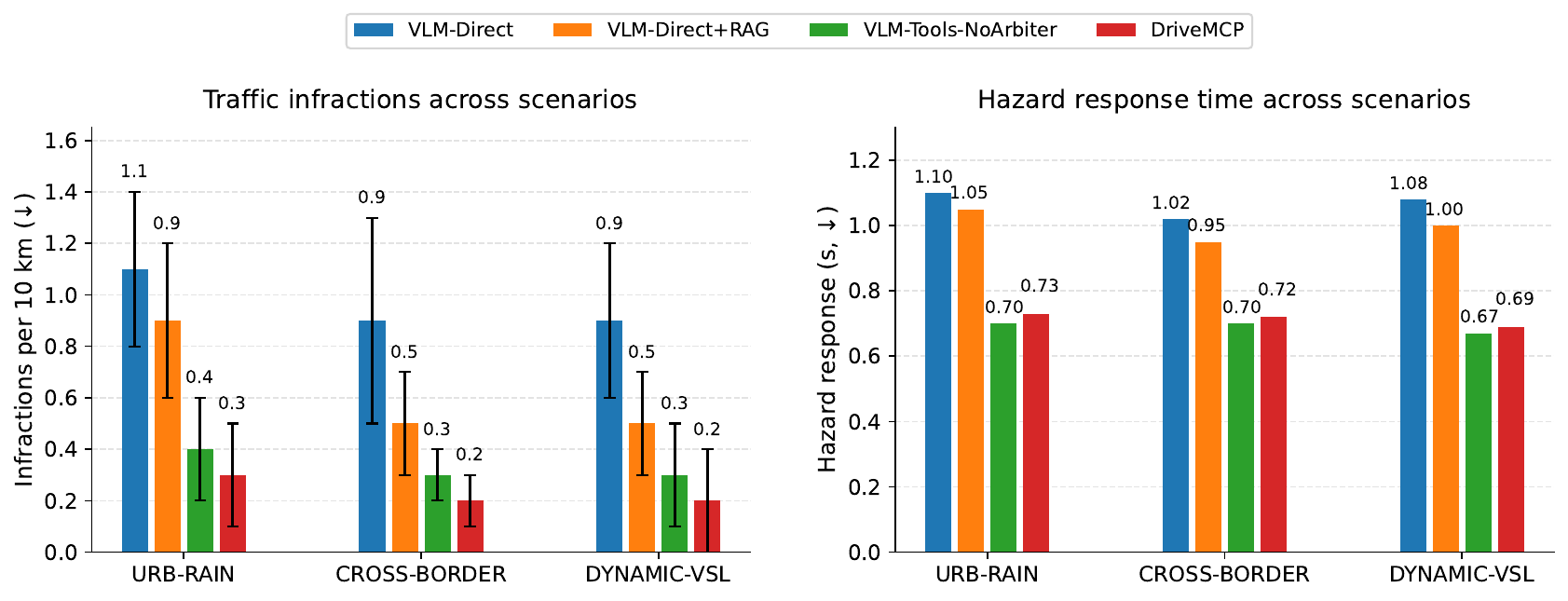}
                                                                                                                                                                                                    \caption{Infractions and hazard response times across scenarios for VLM-Direct, VLM-Direct+RAG, VLM-Tools-NoArbiter, and DriveMCP.}
                                                                                                                                                                                                    \label{fig:bars}
                                                                                                                                                                                                    \end{figure*}

                                                                                                                                                                                                    \subsection{Robustness to Corrupted World-State Fields}
                                                                                                                                                                                                    Table~\ref{tab:corruption} quantifies how perception corruption affects event extraction, tool routing, and downstream safety and compliance.
                                                                                                                                                                                                    Missed or misread speed-limit cues primarily degrade compliance (infractions and overspeed), while actor dropouts primarily degrade hazard response.

                                                                                                                                                                                                    \begin{table*}[t]
                                                                                                                                                                                                    \centering
                                                                                                                                                                                                    \caption{Robustness under controlled \texttt{world\_state} corruption in CROSS-BORDER (DriveMCP). Event F1 and correct-tool rate measure event extraction and routing quality.}
                                                                                                                                                                                                    \label{tab:corruption}
                                                                                                                                                                                                    \setlength{\tabcolsep}{5pt}
                                                                                                                                                                                                    \begin{tabular*}{\textwidth}{@{\extracolsep{\fill}}lccccc}
                                                                                                                                                                                                    \toprule
                                                                                                                                                                                                    Corruption setting &
                                                                                                                                                                                                    \makecell{Event F1 (\%)\\$\uparrow$} &
                                                                                                                                                                                                    \makecell{Correct-tool rate (\%)\\$\uparrow$} &
                                                                                                                                                                                                    \makecell{Infractions/10 km\\$\downarrow$} &
                                                                                                                                                                                                    \makecell{Overspeed (\%)\\$\downarrow$} &
                                                                                                                                                                                                    \makecell{Hazard response (s)\\$\downarrow$} \\
                                                                                                                                                                                                    \midrule
                                                                                                                                                                                                    Clean sensor stack & 92 & 94 & $0.2\pm0.1$ & 0.7 & 0.72 \\
                                                                                                                                                                                                    Sign miss (10\%) & 85 & 88 & $0.4\pm0.2$ & 1.0 & 0.75 \\
                                                                                                                                                                                                    Sign miss (30\%) & 70 & 76 & $0.8\pm0.3$ & 1.8 & 0.80 \\
                                                                                                                                                                                                    Limit misread (10\%) & 78 & 82 & $0.6\pm0.2$ & 1.5 & 0.76 \\
                                                                                                                                                                                                    Actor dropout (10\%) & 90 & 92 & $0.3\pm0.1$ & 0.8 & 0.88 \\
                                                                                                                                                                                                    Actor dropout (30\%) & 84 & 86 & $0.5\pm0.2$ & 1.0 & 1.05 \\
                                                                                                                                                                                                    \bottomrule
                                                                                                                                                                                                    \end{tabular*}
                                                                                                                                                                                                    \end{table*}

                                                                                                                                                                                                    \subsection{Safety Arbiter Calibration and Sensitivity}
                                                                                                                                                                                                    All safety-arbiter constants are fixed across scenarios after offline calibration.
                                                                                                                                                                                                    To assess sensitivity, we vary the speak-versus-act margin $\Delta$ while holding all other parameters constant.
                                                                                                                                                                                                    Table~\ref{tab:delta} shows the expected trade-off: increasing $\Delta$ reduces intervention frequency and false positives but can delay escalation, reducing TTC at escalation.

                                                                                                                                                                                                    \begin{table}[t]
                                                                                                                                                                                                    \centering
                                                                                                                                                                                                    \caption{Sensitivity to arbitration margin $\Delta$ in DYNAMIC-VSL (DriveMCP).}
                                                                                                                                                                                                    \label{tab:delta}
                                                                                                                                                                                                    \setlength{\tabcolsep}{4pt}
                                                                                                                                                                                                    \begin{tabular}{cccc}
                                                                                                                                                                                                    \toprule
                                                                                                                                                                                                    $\Delta$ &
                                                                                                                                                                                                    \makecell{Interventions/10 km\\$\downarrow$} &
                                                                                                                                                                                                    \makecell{False positives (\%)\\$\downarrow$} &
                                                                                                                                                                                                    \makecell{TTC at escalation (s)\\$\uparrow$} \\
                                                                                                                                                                                                    \midrule
                                                                                                                                                                                                    0.0 & 3.2 & 18 & 2.3 \\
                                                                                                                                                                                                    0.1 & 2.4 & 12 & 2.2 \\
                                                                                                                                                                                                    0.2 & 1.8 & 8 & 2.1 \\
                                                                                                                                                                                                    0.3 & 1.2 & 5 & 2.0 \\
                                                                                                                                                                                                    0.4 & 0.9 & 4 & 1.9 \\
                                                                                                                                                                                                    \bottomrule
                                                                                                                                                                                                    \end{tabular}
                                                                                                                                                                                                    \end{table}

                                                                                                                                                                                                    \subsection{Ablation Study}
                                                                                                                                                                                                    Table~\ref{tab:ablation} reports ablations with exact values and 95\% confidence intervals (CI) over 10 runs in CROSS-BORDER.
                                                                                                                                                                                                    Removing retrieval from the Rules server is the dominant contributor to infractions under jurisdiction shifts.
                                                                                                                                                                                                    Removing tool parallelism primarily impacts latency.
                                                                                                                                                                                                    The oracle variant quantifies the remaining gap between DriveLM-derived key facts and simulator metadata.

                                                                                                                                                                                                    \begin{table*}[t]
                                                                                                                                                                                                    \centering
                                                                                                                                                                                                    \caption{Ablations in CROSS-BORDER (mean over 10 runs, with 95\% CI for the mean).}
                                                                                                                                                                                                    \label{tab:ablation}
                                                                                                                                                                                                    \setlength{\tabcolsep}{5pt}
                                                                                                                                                                                                    \begin{tabular*}{\textwidth}{@{\extracolsep{\fill}}lccc}
                                                                                                                                                                                                    \toprule
                                                                                                                                                                                                    Variant &
                                                                                                                                                                                                    \makecell{Infractions/10 km\\$\downarrow$} &
                                                                                                                                                                                                    \makecell{Advisory latency (s)\\$\downarrow$} &
                                                                                                                                                                                                    \makecell{Hazard response (s)\\$\downarrow$} \\
                                                                                                                                                                                                    \midrule
                                                                                                                                                                                                    DriveMCP (full) & $0.20\pm0.06$ & $0.31\pm0.03$ & 0.72 \\
                                                                                                                                                                                                    No RAG in Rules & $0.90\pm0.19$ & $0.32\pm0.03$ & 0.74 \\
                                                                                                                                                                                                    No tool parallelism & $0.24\pm0.07$ & $0.42\pm0.04$ & 0.73 \\
                                                                                                                                                                                                    No CAN context & $0.22\pm0.06$ & $0.31\pm0.03$ & 0.72 \\
                                                                                                                                                                                                    No feedback adaptation & $0.26\pm0.07$ & $0.31\pm0.03$ & 0.75 \\
                                                                                                                                                                                                    Oracle key facts (metadata) & $0.15\pm0.05$ & $0.30\pm0.03$ & 0.71 \\
                                                                                                                                                                                                    \bottomrule
                                                                                                                                                                                                    \end{tabular*}
                                                                                                                                                                                                    \end{table*}

                                                                                                                                                                                                    \subsection{CAN Fault Injection Experiment}
                                                                                                                                                                                                    To evaluate CAN-aware assistance, we inject degraded braking envelopes and corresponding CAN fault indicators (ABS fault and low tire pressure) while keeping the visual scene unchanged.
                                                                                                                                                                                                    Table~\ref{tab:canexp} shows that CAN-aware context reduces recommended target speeds and improves TTC at escalation, which reduces near-collisions in the degraded braking setting.

                                                                                                                                                                                                    \begin{table}[t]
                                                                                                                                                                                                    \centering
                                                                                                                                                                                                    \caption{CAN fault injection scenario (degraded braking). Lower is better for near-collisions; higher is better for TTC.}
                                                                                                                                                                                                    \label{tab:canexp}
                                                                                                                                                                                                    \setlength{\tabcolsep}{3pt}
                                                                                                                                                                                                    \footnotesize
                                                                                                                                                                                                    \begin{tabular}{p{0.23\columnwidth}ccc}
                                                                                                                                                                                                    \toprule
                                                                                                                                                                                                    Method &
                                                                                                                                                                                                    \makecell{Target (km/h)\\$\downarrow$} &
                                                                                                                                                                                                    \makecell{TTC (s)\\$\uparrow$} &
                                                                                                                                                                                                    \makecell{Near-collisions/10 km\\$\downarrow$} \\
                                                                                                                                                                                                    \midrule
                                                                                                                                                                                                    VLM-Direct & 78 & 1.5 & 0.9 \\
                                                                                                                                                                                                    VLM-Direct+RAG & 77 & 1.6 & 0.8 \\
                                                                                                                                                                                                    VLM-Tools-NoArbiter & 70 & 2.0 & 0.4 \\
                                                                                                                                                                                                    DriveMCP (ours) & 65 & 2.2 & 0.2 \\
                                                                                                                                                                                                    \bottomrule
                                                                                                                                                                                                    \end{tabular}
                                                                                                                                                                                                    \end{table}

                                                                                                                                                                                                    \subsection{Qualitative Observations}
                                                                                                                                                                                                    In static regulation scenes, DriveMCP explanations closely follow retrieved statute text, improving transparency compared to monolithic responses.
                                                                                                                                                                                                    In URB-RAIN-MULTILINGUAL, the Weather server contributes traction-aware guidance (reduced target speed and increased headway) that the Rules server alone would not infer.
                                                                                                                                                                                                    CAN context is most useful when the driver asks diagnostic questions (e.g., warning lights) or when the health score indicates degraded braking, in which case the system adopts more conservative advice while preserving legal compliance.

                                                                                                                                                                                                    \section{Discussion: Traceability, Human Factors, and CAN Use Cases}
                                                                                                                                                                                                    \subsection{Traceability and Post-Drive Audit}
                                                                                                                                                                                                    A core motivation for tool modularity is the ability to reconstruct \emph{why} a recommendation was produced.
                                                                                                                                                                                                    DriveMCP records (i) the detected events, (ii) the exact set of MCP tools invoked, (iii) the retrieved rule passages (document identifiers and offsets), (iv) intermediate tool outputs, (v) the fused explanation, and (vi) the safety metrics and arbitration outcome.
                                                                                                                                                                                                    This trace supports debugging and regression testing, policy review, and model update containment.

                                                                                                                                                                                                    \subsection{Human Factors and Explanation Quality}
                                                                                                                                                                                                    Driver assistance must balance informativeness with cognitive load.
                                                                                                                                                                                                    DriveMCP controls verbosity using a two-level explanation:
                                                                                                                                                                                                    \emph{(i)} a short actionable directive (``slow to 60 km/h''), and \emph{(ii)} an optional justification (``gantry limit updated; heavy rain reduces traction'').
                                                                                                                                                                                                    The orchestrator suppresses repeated messages using short-term memory and adjusts tone based on feedback labels.
                                                                                                                                                                                                    This design aligns with prior work on self-driving explanations that emphasizes concise, user-oriented rationales \cite{kim_eccv_2018}.

                                                                                                                                                                                                    \subsection{CAN-Based Diagnostic and Degraded-Mode Assistance}
                                                                                                                                                                                                    CAN/OBD context enables assistance beyond vision-only interpretation.
                                                                                                                                                                                                    For example, if a stability-control warning is present, the assistant can recommend a lower speed and smoother braking, even when the scene appears benign.
                                                                                                                                                                                                    Likewise, a low tire pressure signal can increase hydroplaning risk under rain and motivate earlier speed reduction.
                                                                                                                                                                                                    In DriveMCP, CAN-derived constraints modify the health score $h$ and are surfaced explicitly to the driver.

                                                                                                                                                                                                    \begin{table}[t]
                                                                                                                                                                                                    \centering
                                                                                                                                                                                                    \caption{Illustrative DriveMCP interactions (abbreviated).}
                                                                                                                                                                                                    \label{tab:examples}
                                                                                                                                                                                                    \setlength{\tabcolsep}{4pt}
                                                                                                                                                                                                    \begin{tabular}{p{0.20\columnwidth}p{0.33\columnwidth}p{0.35\columnwidth}}
                                                                                                                                                                                                    \toprule
                                                                                                                                                                                                    Scenario & Driver query & Tools invoked $\rightarrow$ key output \\
                                                                                                                                                                                                    \midrule
                                                                                                                                                                                                    URB-RAIN & ``What does this sign mean?'' & Rules + Weather $\rightarrow$ stop required; reduce speed due to low friction.\\
                                                                                                                                                                                                    CROSS-BORDER & ``Are we in mph now?'' & Rules $\rightarrow$ jurisdiction switched; units converted; new limit explained.\\
                                                                                                                                                                                                    DYNAMIC-VSL & ``Why did the limit drop?'' & Rules + Weather $\rightarrow$ gantry update; recommend compliant and safe target speed.\\
                                                                                                                                                                                                    CAN fault & ``A warning light came on; is it safe?'' & CAN + Rules $\rightarrow$ DTC/health context; recommend conservative driving or safe pull-over.\\
                                                                                                                                                                                                    \bottomrule
                                                                                                                                                                                                    \end{tabular}
                                                                                                                                                                                                    \end{table}

                                                                                                                                                                                                    \section{Limitations and Future Work}
                                                                                                                                                                                                    \subsection{Perception and DriveLM Integration}
                                                                                                                                                                                                    DriveMCP is evaluated in simulation using a sensor-like stack that derives \texttt{world\_state} from camera and LiDAR perception, with key compliance fields parsed from DriveLM outputs.
                                                                                                                                                                                                    However, important gaps remain relative to on-road deployment, including sensor calibration errors, domain shift, and temporal inconsistency in VLM outputs under extreme conditions.
                                                                                                                                                                                                    Future work should evaluate multi-camera inputs, richer perception stacks, and stronger uncertainty propagation, including conservative strategies for uncertain sign and jurisdiction recognition.

                                                                                                                                                                                                    \subsection{Knowledge Quality and Jurisdictional Coverage}
                                                                                                                         DriveMCP relies on external corpora for compliance reasoning.
                                                                                                                                                                                                    Although this improves updateability, it introduces new failure modes: stale rule packs, incomplete jurisdiction coverage, and retrieval errors.
                                                                                                                                                                                                    Future work should formalize corpus curation and versioning, incorporate automated consistency checks, and add confidence estimation for retrieval.

                                                                                                                                                                                                    \subsection{Tool Reliability, Security, and Privacy}
                                                                                                                                                                                                    Modular MCP servers simplify system composition but require robust security.
                                                                                                                                                                                                    A production system would need authenticated tool endpoints, access control, and sandboxing to prevent prompt injection or malicious tool outputs.
                                                                                                                                                                                                    For CAN access, privacy concerns require minimizing signal requests and restricting logs to safety-relevant fields.

                                                                                                                                                                                                    \subsection{Safety Assurance and Certification Pathways}
                                                                                                                                                                                                    The RSS/TTC arbiter in DriveMCP is intentionally simplified and should not be interpreted as a certified safety controller.
                                                                                                                                                                                                    We view DriveMCP as a research prototype that demonstrates how interpretable guardrails and auditable tool traces can support safety case arguments, but full assurance remains an open challenge.
                                                                                                                                                                                                    Relevant safety standards include ISO 26262 and ISO 21448 \cite{iso26262, iso21448}.

                                                                                                                                                                                                    \subsection{Human Factors and Overreliance}
                                                                                                                                                                                                    Language-based assistants can shape driver behavior.
                                                                                                                                                                                                    Future work should measure driver trust calibration, distraction, and potential overreliance under long drives, including message frequency and phrasing.

                                                                                                                                                                                                    \subsection{Future Extensions}
                                                                                                                                                                                                    Beyond the three MCP servers used in this work, additional domain services could include road-work feeds, HD map updates, and fleet learning signals.
                                                                                                                                                                                                    Multi-agent coordination perspectives (e.g., swarm intelligence) may also inform cooperative driving and fleet-scale adaptation \cite{nadiri_sensors_swarm_2025}.
                                                                                                                                                                                                    We also plan to evaluate DriveMCP against closed-loop benchmarks and to release reproducible tool-call traces and scenario specifications to support comparative research.

                                                                                                                                                                                                    \section{Conclusion}
                                                                                                                                                                                                    This paper presented DriveMCP, a DriveLM-grounded, MCP-orchestrated driver-assistance framework that combines regulation-aware retrieval, weather-aware traction reasoning, CAN-based vehicle context, and RSS-inspired safety arbitration.
                                                                                                                                                                                                    By modularizing perception, compliance, environment reasoning, and telemetry interpretation, DriveMCP improves updateability and traceability while retaining low-latency responsiveness through parallel tool calls.
                                                                                                                                                                                                    Simulation results demonstrate reduced infractions and improved hazard response relative to VLM-Direct, VLM-Direct+RAG, and VLM-Tools-NoArbiter baselines.
                                                                                                                                                                                                    The proposed architecture provides a practical template for integrating VLM-based understanding with vehicle-grade constraints and auditable decision logic in intelligent vehicles.

                                                                                                                                                                                                                                                                                                                                                                                                                          % Biographies are typically required at final submission for many IEEE journals.

\begin{thebibliography}{99}

                                                                                                                                                                                                                                                                                                                                                                                                                          \bibitem{paden_tiv_2016}
                                                                                                                                                                                                                                                                                                                                                                                                                          B.~Paden, M.~\v{C}\'{a}p, S.~Z.~Yong, D.~Yershov, and E.~Frazzoli, ``A survey of motion planning and control techniques for self-driving urban vehicles,'' \emph{IEEE Trans.\ Intell.\ Veh.}, vol.~1, no.~1, pp.~33--55, Mar.~2016, doi: 10.1109/TIV.2016.2578706.

                                                                                                                                                                                                                                                                                                                                                                                                                          \bibitem{grigorescu_jfr_2020}
                                                                                                                                                                                                                                                                                                                                                                                                                          S.~Grigorescu, B.~Trasnea, T.~Cocias, and G.~Macesanu, ``A survey of deep learning techniques for autonomous driving,'' \emph{J.\ Field Robot.}, vol.~37, no.~3, pp.~362--386, Apr.~2020, doi: 10.1002/rob.21918.

                                                                                                                                                                                                                                                                                                                                                                                                                          \bibitem{chen_tpami_2024}
                                                                                                                                                                                                                                                                                                                                                                                                                          L.~Chen, P.~Wu, K.~Chitta, B.~Jaeger, A.~Geiger, and H.~Li, ``End-to-end autonomous driving: Challenges and frontiers,'' \emph{IEEE Trans.\ Pattern Anal.\ Mach.\ Intell.}, early access, 2024, doi: 10.1109/TPAMI.2024.3435937.

                                                                                                                                                                                                                                                                                                                                                                                                                          \bibitem{drivegpt4}
                                                                                                                                                                                                                                                                                                                                                                                                                          Z.~Xu, Y.~Zhang, E.~Xie, Z.~Zhao, Y.~Guo, K.-Y.~K.~Wong, Z.~Li, and H.~Zhao, ``DriveGPT4: Interpretable end-to-end autonomous driving via large language model,'' \emph{IEEE Robot.\ Autom.\ Lett.}, vol.~9, no.~10, pp.~8186--8193, 2024, doi: 10.1109/LRA.2024.3440097.

                                                                                                                                                                                                                                                                                                                                                                                                                          \bibitem{drivelm}
                                                                                                                                                                                                                                                                                                                                                                                                                          C.~Sima \emph{et al.}, ``DriveLM: Driving with graph visual question answering,'' in \emph{Proc.\ Eur.\ Conf.\ Comput.\ Vis.\ (ECCV)}, 2024, pp.~261--279, doi: 10.1007/978-3-031-72943-0\_15.

                                                                                                                                                                                                                                                                                                                                                                                                                          \bibitem{vlp}
                                                                                                                                                                                                                                                                                                                                                                                                                          C.~Pan, B.~Yaman, T.~Nesti, A.~Mallik, A.~G.~Allievi, S.~Velipasalar, and L.~Ren, ``VLP: Vision language planning for autonomous driving,'' in \emph{Proc.\ IEEE/CVF Conf.\ Comput.\ Vis.\ Pattern Recognit.\ (CVPR)}, 2024, pp.~14760--14769.

                                                                                                                                                                                                                                                                                                                                                                                                                          \bibitem{vlaad}
                                                                                                                                                                                                                                                                                                                                                                                                                          S.~Park, M.~Lee, J.~Kang, H.~Choi, Y.~Park, J.~Cho, A.~Lee, and D.~Kim, ``VLAAD: Vision and language assistant for autonomous driving,'' in \emph{Proc.\ IEEE/CVF Winter Conf.\ Appl.\ Comput.\ Vis.\ Workshops (WACVW)}, 2024, pp.~980--987.

                                                                                                                                                                                                                                                                                                                                                                                                                          \bibitem{simlingo}
                                                                                                                                                                                                                                                                                                                                                                                                                          K.~Renz, L.~Chen, E.~Arani, and O.~Sinavski, ``SimLingo: Vision-only closed-loop autonomous driving with language-action alignment,'' arXiv:2503.09594, 2025. [Online]. Available: \url{https://arxiv.org/abs/2503.09594}

                                                                                                                                                                                                                                                                                                                                                                                                                          \bibitem{vla_survey}
                                                                                                                                                                                                                                                                                                                                                                                                                          S.~Jiang \emph{et al.}, ``A survey on vision-language-action models for autonomous driving,'' arXiv:2506.24044, 2025. [Online]. Available: \url{https://arxiv.org/abs/2506.24044}

                                                                                                                                                                                                                                                                                                                                                                                                                          \bibitem{ragdriver}
                                                                                                                                                                                                                                                                                                                                                                                                                          J.~Yuan \emph{et al.}, ``RAG-Driver: Generalisable driving explanations with retrieval-augmented in-context learning in multi-modal large language model,'' arXiv:2402.10828, 2024. [Online]. Available: \url{https://arxiv.org/abs/2402.10828}

                                                                                                                                                                                                                                                                                                                                                                                                                          \bibitem{kim_eccv_2018}
                                                                                                                                                                                                                                                                                                                                                                                                                          J.~Kim, A.~Rohrbach, T.~Darrell, and J.~Canny, ``Textual explanations for self-driving vehicles,'' in \emph{Proc.\ Eur.\ Conf.\ Comput.\ Vis.\ (ECCV)}, 2018, pp.~563--578, doi: 10.1007/978-3-030-01216-8\_35.

                                                                                                                                                                                                                                                                                                                                                                                                                          \bibitem{rag}
                                                                                                                                                                                                                                                                                                                                                                                                                          P.~Lewis \emph{et al.}, ``Retrieval-augmented generation for knowledge-intensive NLP tasks,'' in \emph{Proc.\ Adv.\ Neural Inf.\ Process.\ Syst.\ (NeurIPS)}, 2020. [Online]. Available: \url{https://arxiv.org/abs/2005.11401}

                                                                                                                                                                                                                                                                                                                                                                                                                          \bibitem{bm25}
                                                                                                                                                                                                                                                                                                                                                                                                                          S.~Robertson and H.~Zaragoza, ``The probabilistic relevance framework: BM25 and beyond,'' \emph{Found.\ Trends Inf.\ Retr.}, vol.~3, no.~4, pp.~333--389, 2009, doi: 10.1561/1500000019.

                                                                                                                                                                                                                                                                                                                                                                                                                          \bibitem{dpr}
                                                                                                                                                                                                                                                                                                                                                                                                                          V.~Karpukhin \emph{et al.}, ``Dense passage retrieval for open-domain question answering,'' arXiv:2004.04906, 2020. [Online]. Available: \url{https://arxiv.org/abs/2004.04906}

                                                                                                                                                                                                                                                                                                                                                                                                                          \bibitem{mcp}
                                                                                                                                                                                                                                                                                                                                                                                                                          Model Context Protocol, ``MCP specification and documentation,'' accessed 2025. [Online]. Available: \url{https://modelcontextprotocol.io/}

                                                                                                                                                                                                                                                                                                                                                                                                                          \bibitem{langgraph}
                                                                                                                                                                                                                                                                                                                                                                                                                          LangChain, ``LangGraph overview: Durable, stateful, human-in-the-loop agent graphs,'' accessed 2025. [Online]. Available: \url{https://docs.langchain.com/oss/python/langgraph/overview}

                                                                                                                                                                                                                                                                                                                                                                                                                          \bibitem{mrkl}
                                                                                                                                                                                                                                                                                                                                                                                                                          A.~Karpas \emph{et al.}, ``MRKL systems: A modular, neuro-symbolic architecture that combines large language models, external knowledge sources and discrete reasoning,'' arXiv:2205.00445, 2022. [Online]. Available: \url{https://arxiv.org/abs/2205.00445}

                                                                                                                                                                                                                                                                                                                                                                                                                          \bibitem{react}
                                                                                                                                                                                                                                                                                                                                                                                                                          S.~Yao \emph{et al.}, ``ReAct: Synergizing reasoning and acting in language models,'' in \emph{Proc.\ Int.\ Conf.\ Learn.\ Represent.\ (ICLR)}, 2023. [Online]. Available: \url{https://arxiv.org/abs/2210.03629}

                                                                                                                                                                                                                                                                                                                                                                                                                          \bibitem{toolformer}
                                                                                                                                                                                                                                                                                                                                                                                                                          T.~Schick \emph{et al.}, ``Toolformer: Language models can teach themselves to use tools,'' in \emph{Proc.\ Adv.\ Neural Inf.\ Process.\ Syst.\ (NeurIPS)}, 2023. [Online]. Available: \url{https://arxiv.org/abs/2302.04761}

                                                                                                                                                                                                                                                                                                                                                                                                                          \bibitem{rss}
                                                                                                                                                                                                                                                                                                                                                                                                                          S.~Shalev-Shwartz, S.~Shammah, and A.~Shashua, ``On a formal model of safe and scalable self-driving cars,'' arXiv:1708.06374, 2017. [Online]. Available: \url{https://arxiv.org/abs/1708.06374}

                                                                                                                                                                                                                                                                                                                                                                                                                          \bibitem{adrsslib}
                                                                                                                                                                                                                                                                                                                                                                                                                          Intel, ``ad-rss-lib: C++ library implementing the Responsibility-Sensitive Safety (RSS) model,'' GitHub repository, accessed 2025. [Online]. Available: \url{https://github.com/intel/ad-rss-lib}

                                                                                                                                                                                                                                                                                                                                                                                                                          \bibitem{hayward_ttc}
                                                                                                                                                                                                                                                                                                                                                                                                                          J.~C.~Hayward, ``Near miss determination through use of a scale of danger,'' \emph{Highway Research Record}, no.~384, pp.~24--34, 1972.

                                                                                                                                                                                                                                                                                                                                                                                                                          \bibitem{rajamani}
                                                                                                                                                                                                                                                                                                                                                                                                                          R.~Rajamani, \emph{Vehicle Dynamics and Control}, 2nd~ed. Springer, 2012.

                                                                                                                                                                                                                                                                                                                                                                                                                          \bibitem{iso26262}
                                                                                                                                                                                                                                                                                                                                                                                                                          ISO, \emph{ISO 26262 Road Vehicles -- Functional Safety}. International Organization for Standardization, 2018.

                                                                                                                                                                                                                                                                                                                                                                                                                          \bibitem{iso21448}
                                                                                                                                                                                                                                                                                                                                                                                                                          ISO, \emph{ISO 21448 Road Vehicles -- Safety of the Intended Functionality (SOTIF)}. International Organization for Standardization, 2022.

                                                                                                                                                                                                                                                                                                                                                                                                                          \bibitem{vienna_convention}
                                                                                                                                                                                                                                                                                                                                                                                                                          United Nations, ``Vienna Convention on Road Signs and Signals,'' 1968. [Online]. Available: \url{https://treaties.un.org/}

                                                                                                                                                                                                                                                                                                                                                                                                                          \bibitem{mutcd}
                                                                                                                                                                                                                                                                                                                                                                                                                          Federal Highway Administration, ``Manual on Uniform Traffic Control Devices (MUTCD), 11th Edition,'' 2023. [Online]. Available: \url{https://mutcd.fhwa.dot.gov/}

                                                                                                                                                                                                                                                                                                                                                                                                                          \bibitem{tsrgd}
                                                                                                                                                                                                                                                                                                                                                                                                                          U.K.\ Government, ``The Traffic Signs Regulations and General Directions 2016,'' 2016. [Online]. Available: \url{https://www.legislation.gov.uk/}

                                                                                                                                                                                                                                                                                                                                                                                                                          \bibitem{carla}
                                                                                                                                                                                                                                                                                                                                                                                                                          A.~Dosovitskiy \emph{et al.}, ``CARLA: An open urban driving simulator,'' in \emph{Proc.\ Conf.\ Robot Learn.\ (CoRL)}, 2017.

                                                                                                                                                                                                                                                                                                                                                                                                                          \bibitem{nuscenes}
                                                                                                                                                                                                                                                                                                                                                                                                                          H.~Caesar \emph{et al.}, ``nuScenes: A multimodal dataset for autonomous driving,'' in \emph{Proc.\ IEEE/CVF Conf.\ Comput.\ Vis.\ Pattern Recognit.\ (CVPR)}, 2020, pp.~11621--11631.

                                                                                                                                                                                                                                                                                                                                                                                                                          \bibitem{yolo}
                                                                                                                                                                                                                                                                                                                                                                                                                          J.~Redmon, S.~Divvala, R.~Girshick, and A.~Farhadi, ``You only look once: Unified, real-time object detection,'' in \emph{Proc.\ IEEE Conf.\ Comput.\ Vis.\ Pattern Recognit.\ (CVPR)}, 2016, pp.~779--788, doi: 10.1109/CVPR.2016.91.

                                                                                                                                                                                                                                                                                                                                                                                                                          \bibitem{kalman}
                                                                                                                                                                                                                                                                                                                                                                                                                          R.~E.~Kalman, ``A new approach to linear filtering and prediction problems,'' \emph{J.\ Basic Eng.}, vol.~82, no.~1, pp.~35--45, 1960, doi: 10.1115/1.3662552.

                                                                                                                                                                                                                                                                                                                                                                                                                          \bibitem{sort}
                                                                                                                                                                                                                                                                                                                                                                                                                          A.~Bewley, Z.~Ge, L.~Ott, F.~Ramos, and B.~Upcroft, ``Simple online and realtime tracking,'' in \emph{Proc.\ IEEE Int.\ Conf.\ Image Process.\ (ICIP)}, 2016, pp.~3464--3468, doi: 10.1109/ICIP.2016.7533003.

                                                                                                                                                                                                                                                                                                                                                                                                                          \bibitem{iso11898}
                                                                                                                                                                                                                                                                                                                                                                                                                          ISO, \emph{ISO 11898-1 Road Vehicles -- Controller Area Network (CAN) -- Part 1: Data Link Layer and Physical Signalling}. International Organization for Standardization, 2015.

                                                                                                                                                                                                                                                                                                                                                                                                                          \bibitem{sae_j1979}
                                                                                                                                                                                                                                                                                                                                                                                                                          SAE International, \emph{SAE J1979: E/E Diagnostic Test Modes}. SAE International, 2014.

                                                                                                                                                                                                                                                                                                                                                                                                                          \bibitem{mcp_can}
                                                                                                                                                                                                                                                                                                                                                                                                                          F.~Nadiri, ``MCP-CAN: Model Context Protocol server for CAN and OBD telemetry,'' GitHub repository, accessed 2025. [Online]. Available: \url{https://github.com/farzadnadiri/MCP-CAN}

                                                                                                                                                                                                                                                                                                                                                                                                                          \bibitem{nadiri_machines_lookdown_2025}
                                                                                                                                                                                                                                                                                                                                                                                                                          F.~Nadiri and A.~B.~Rad, ``A novel lateral control system for autonomous vehicles: A look-down strategy,'' \emph{Machines}, vol.~13, no.~3, p.~211, Mar.~2025, doi: 10.3390/machines13030211.

                                                                                                                                                                                                                                                                                                                                                                                                                          \bibitem{nadiri_ijira_localization_2025}
                                                                                                                                                                                                                                                                                                                                                                                                                          F.~Nadiri, T.~Banirostam, and A.~B.~Rad, ``On a novel localization methodology for humanoid soccer robots via sensor fusion and perspective transformation,'' \emph{Int.\ J.\ Intell.\ Robot.\ Appl.}, vol.~9, pp.~1577--1593, Apr.~2025, doi: 10.1007/s41315-025-00451-5.

                                                                                                                                                                                                                                                                                                                                                                                                                          \bibitem{nadiri_sensors_swarm_2025}
                                                                                                                                                                                                                                                                                                                                                                                                                          F.~Nadiri and A.~B.~Rad, ``Swarm intelligence for collaborative play in humanoid soccer teams,'' \emph{Sensors}, vol.~25, no.~11, p.~3496, May~2025, doi: 10.3390/s25113496.

                                                                                                                                                                                                                                                                                                                                                                                                                          \end{thebibliography}
                                                                                                                                                                                                                                                                                                                                                                                                                       \end{document}